\pdfoutput=1

\documentclass[11pt]{article}

\usepackage[final]{acl}

\usepackage{times}
\usepackage{latexsym}

\usepackage[T1]{fontenc}

\usepackage[utf8]{inputenc}

\usepackage{microtype}

\usepackage{inconsolata}

\usepackage{graphicx}

\usepackage[whole]{bxcjkjatype}
\usepackage{lingmacros}
\usepackage{amsmath,amssymb}
\usepackage{mathtools}
\usepackage{ascmac}
\usepackage{proof,lscape,caption}
\usepackage{multirow}
\usepackage{color}
\usepackage{stmaryrd}
\usepackage{url}
\usepackage{booktabs}
\usepackage{natbib}
\bibpunct[:]{(}{)}{,}{a}{}{,}
\usepackage{pifont}
\usepackage{enumerate}
\usepackage[margin=5pt]{subcaption}
\usepackage{makecell}
\usepackage{bussproofs}
\usepackage{lscape}
\usepackage{bm}
\usepackage{longtable}
\usepackage{leipzig}
\usepackage{gb4e}
\noautomath
\usepackage{breqn}
\usepackage{cleveref}
\usepackage{here}

\newcommand{\AXC}[1]{\AxiomC{#1}}
\newcommand{\UIC}[1]{\UnaryInfC{#1}}
\newcommand{\BIC}[1]{\BinaryInfC{#1}}

\newcommand{\RL}[1]{\RightLabel{#1}}

    {\gdef\scalefactor{#1}\begin{center}\proofSkipAmount \leavevmode}%
    {\scalebox{\scalefactor}{\DisplayProof}\proofSkipAmount \end{center}%
}

\newcommand{\fcf}{{>}\mathbf{B}}
\newcommand{\fcb}{{<}\mathbf{B}}
\newcommand{\cfcf}{{>}\mathbf{B}_{\times}}

\newcommand{\catLF}[2]{{#1}:{#2}}
\newcommand{\catLFbr}[2]{\begin{matrix}{#1}\\{{}:{#2}}\end{matrix}}

\makeglossaries

\newleipzig{npst}{npst}{nonpast}
\newleipzig{qp}{qp}{questionparticle}
\newleipzig{no}{no}{nominalizer}
\newcommand{\bs}{\backslash}
\newcommand{\NP}{\mathit{NP}}
\newcommand{\heavy}{\mathsf{heavy}}
\newcommand{\light}{\mathsf{light}}
\newcommand{\tall}{\mathsf{tall}}

\newcommand{\john}{\mathsf{john}}

\newcommand{\bob}{\mathsf{bob}}
\newcommand{\taro}{\mathsf{taro}}
\newcommand{\jiro}{\mathsf{jiro}}
\newcommand{\hanako}{\mathsf{hanako}}

\newcommand{\book}{\mathsf{book}}
\newcommand{\nom}{\mathsf{Nom}}
\newcommand{\acc}{\mathsf{Acc}}
\newcommand{\student}{\mathsf{student}}
\newcommand{\expensive}{\mathsf{expensive}}

\newcommand{\bought}{\mathsf{bought}}

\newcommand{\ran}{\mathsf{ran}}
\newcommand{\know}{\mathsf{know}}
\newcommand{\bA}{\mathbf{A}}
\newcommand{\bP}{\mathbf{P}}
\newcommand{\bN}{\mathbf{N}}

\title{Implementing a Logical Inference System\\ for Japanese Comparatives}

\author{
    Yosuke Mikami${}^{1, 2}$ \quad
    Daiki Matsuoka${}^{1, 2}$ \quad
    Hitomi Yanaka${}^{1, 2}$ \\
    ${}^1$The University of Tokyo\\
    ${}^2$Riken \\
    \texttt{\{ymikami, daiki.matsuoka, hyanaka\}@is.s.u-tokyo.ac.jp}
}

\begin{document}
\maketitle
\begin{abstract}
Natural Language Inference (NLI) involving comparatives is challenging because it requires understanding quantities and comparative relations expressed by sentences.
While some approaches leverage Large Language Models (LLMs), we focus on logic-based approaches grounded in compositional semantics, which are promising for robust handling of numerical and logical expressions.
Previous studies along these lines have proposed logical inference systems for English comparatives.
However, it has been pointed out that there are several morphological and semantic differences between Japanese and English comparatives.
These differences make it difficult to apply such systems directly to Japanese comparatives.
To address this gap, this study proposes ccg-jcomp, a logical inference system for Japanese comparatives based on compositional semantics.
We evaluate the proposed system on a Japanese NLI dataset containing comparative expressions.
We demonstrate the effectiveness of our system by comparing its accuracy with that of existing LLMs.
\end{abstract}

\section{Introduction}
Natural Language Inference (NLI)~\citep{bowman-etal-2015-large} is the task of determining the entailment relation between premise and hypothesis sentences.
In particular, this paper focuses on inferences involving comparative expressions (e.g., \textit{heavier}, where the comparative morpheme \textit{-er} is attached).
In (\ref{comp-example}), for example, the premise (\ref{comp-example-a}) and (\ref{comp-example-b}) entail the hypothesis (\ref{comp-example-c}).
\begin{exe}
    \ex\label{comp-example}
    \begin{xlist}
        \ex\label{comp-example-a}
        John is heavier than Bob.
        \ex\label{comp-example-b}
        Bob is heavier than 70 kg.
        \ex\label{comp-example-c}
        John is heavier than 70 kg.
        \sn\hfill (entailment)
    \end{xlist}
\end{exe}
Inferences involving comparatives like (\ref{comp-example}) are challenging to an NLI system because the system needs to correctly understand the meaning of the quantity expression ``70 kg'' and the comparative relation between John's and Bob's weights.

There are two main approaches to NLI.
One is a deep learning (DL)-based approach.
Large Language Models (LLMs), such as GPT-4o,\footnote{\url{https://openai.com/index/gpt-4o-system-card/}} have been performing accurately in various tasks, including NLI.
However, recent works~\cite{she2023scone,liu2023evaluatinglogicalreasoningability,parmar-etal-2024-logicbench} have pointed out that even such models have difficulties in handling problems involving logical connectives such as negation and quantification.
This fact indicates that DL-based models still have room for improvement.

The other approach to NLI is a logic-based approach~\cite{abzianidze-2015-tableau,mineshima2015higher,bernardy-chatzikyriakidis-2017-type,hu-etal-2020-monalog, bernardy-chatzikyriakidis-2021-applied}, in which mathematical logic is utilized to perform NLI involving various logical expressions robustly.
In particular, inference systems based on compositional semantics have achieved high performance on NLI problems composed of lexical, syntactic, and semantic phenomena.
As for comparatives, \citet{Haruta_Mineshima_Bekki_2022} proposed a logical inference system for English comparatives based on \textit{Combinatory Categorial Grammar} (CCG,~\citealt{steedman2000syntactic}) and \textit{degree semantics}~\citep{cresswell1976semantics,klein1980semantics}.
However, we cannot apply the system directly to Japanese comparatives because of morphological and semantic differences between Japanese and English comparatives, which we will describe in detail in \Cref{sec:semantic-composition}.

\begin{table*}[t]
    \centering
    \begin{tabular}{ll}
    \hline
        Sentence & Semantic Representation \\
    \hline
        John is heavy. & $\heavy(\john, \theta)$ \\
        John is heavier than 70 kg. & $\exists d. \left( \heavy(\john, d) \land d > 70\text{kg} \right)$ \\
        John is heavier than all the student. & $\forall x. (
        \student(x) \rightarrow \exists d. (\heavy(\john, d) \land \lnot \heavy(x, d)))$ \\
    \hline
    \end{tabular}
    \caption{Basic semantic representations for comparatives}
    \label{tab:basic-SR}
\end{table*}

\begin{figure*}[h]
    \centering
    \includegraphics[width=0.9\linewidth]{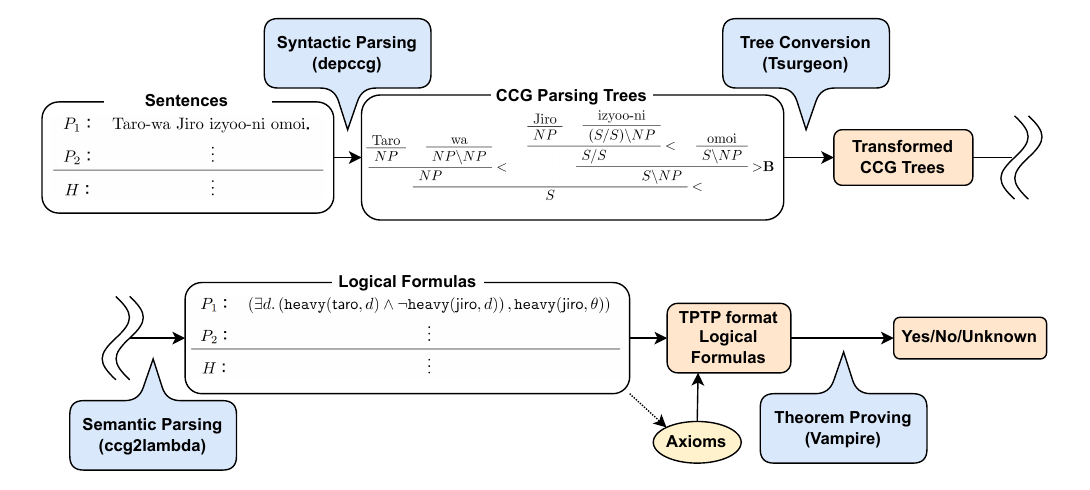}
    \caption{Overview of the proposed system}
    \label{fig:overview}
\end{figure*}

In this study, we aim to develop a logical inference system for Japanese comparatives based on CCG and degree semantics.
Inspired by the logical inference system for English comparatives proposed by~\citet{Haruta_Mineshima_Bekki_2022}, our system, named ccg-jcomp, compositionally derives the semantic representations (i.e., the logical formulas representing the sentence meanings) of Japanese sentences through syntactic and semantic parsing and judges the entailment relation using a theorem prover.
Further, we implement syntactic and semantic analyses to systematically handle some phenomena specific to Japanese comparatives.

We experiment with JSeM~\citep{10.1007/978-3-319-50953-2_5}, a Japanese NLI dataset containing problems involving comparatives.
We compare the performance of our system with GPT-4o and some Japanese LLMs.
Our experiment shows that our proposed system outperforms all of them in accuracy on the dataset.

Our contributions are as follows:
\begin{enumerate}
  \item We compositionally derive the semantic representations of Japanese sentences containing some comparative expressions based on CCG and degree semantics.
  \item We implement ccg-jcomp, a logical inference system for Japanese comparatives.\footnote{Our system is available for research use at \url{https://github.com/ynklab/ccg-jcomp}}
  \item We demonstrate the effectiveness of our proposed system through experiments on a Japanese NLI dataset involving comparatives.
\end{enumerate}

\section{Degree Semantics}

In our study, we adopt a theoretical framework called degree semantics, which allows us to analyze the meanings of gradable adjectives and comparatives formally.
Its basic idea is to treat a gradable adjective as a binary predicate that takes an entity and a degree as arguments.
For instance, ``John is $d$ feet tall'' can be represented as $\tall(\john, d)$ (for simplicity, we omit units such as ``feet'').

We handle comparatives following the so-called A-not-A analysis~\citep{Seuren1973,klein1982interpretation} in degree semantics.
According to this analysis, (\ref{comparatives-a-not-a-1}) can be represented as (\ref{comparatives-a-not-a-2}), which means that there exists a degree $d$ such that John's weight is more than or equal to $d$ and Bob's weight is not.
\begin{exe}
  \ex\label{ex:comparatives-a-not-a}
  \begin{xlist}
    \ex\label{comparatives-a-not-a-1}
    John is heavier than Bob.
    \ex\label{comparatives-a-not-a-2}
    $\exists d. \left( \heavy(\john, d) \land \lnot \heavy(\bob, d) \right)$
  \end{xlist}
\end{exe}

\noindent
\Cref{tab:basic-SR} shows some other examples of basic constructions involving comparatives and their semantic representations.

\section{System Overview}
\label{sec:proposed-system}

\begin{table*}[h]
  \centering
  \begin{tabular}{lll}
    \hline
    Category & Word Type & Semantic Template\\
    \hline \hline
    $\NP$ &
    \begin{tabular}[c]{@{}l@{}}
      common \\
      noun
    \end{tabular}
    & $\lambda E\ N\ F. \exists x.\left( N(E, x) \land F(x) \right)$\\
    \hline
    $S \bs \NP$ &
    \begin{tabular}[c]{@{}l@{}}
      positive \\
      adjective
    \end{tabular} & 
    \begin{tabular}[c]{@{}l@{}}
      $\lambda E\ Q\ N.$
      $Q(\lambda I. I, \lambda x. N(E, \lambda d. d, \lambda d. d, \lambda t. t, x))$
    \end{tabular}\\
    \hline
    $S \bs \NP$ &
    \begin{tabular}[c]{@{}l@{}}
      negative \\
      adjective
    \end{tabular} &
    \begin{tabular}[c]{@{}l@{}}
      $\lambda E\ Q\ N.Q(\lambda I. I, \lambda x. N(E, \lambda d. d, \lambda d. {-d}, \lambda t. \lnot t, x))$
    \end{tabular}\\
    \hline
    $(S / S) \bs \NP$ & yori &
    \begin{tabular}[c]{@{}l@{}}
      $\lambda E\ Q\ V. V(\lambda A\ x.Q(\lambda I. I,$
      $\lambda y. \exists d. (A(x, d) \land \lnot A(y, d))))$
    \end{tabular}\\
    \hline
    $(S / S) \bs \NP$ &
    \begin{tabular}[c]{@{}l@{}}
      yori \\
      (measure \\
      phrase)
    \end{tabular} &
    \begin{tabular}[c]{@{}l@{}}
      $\lambda E\ Q\ V. V(\lambda A\ F\ x.Q(\lambda I. I,$
      $\lambda y. \exists d. (A(x, d) \land F(y < d))))$
    \end{tabular}\\
    \hline
  \end{tabular}
  \caption{Examples of basic semantic templates}
  \label{tab:semantic_templates}
\end{table*}

Figure \ref{fig:overview} shows the overview of ccg-jcomp, our proposed system.
The overall system flow follows \citet{Haruta_Mineshima_Bekki_2022}: CCG syntactic parsing, tree conversion, semantic parsing, and theorem proving.
In what follows, we describe the details of each step, deferring the explanation of the specifics of Japanese comparatives until~\cref{sec:semantic-composition}.

\subsection{Syntactic Parsing}
\label{subsec:syntactic-parsing}

First, a tokenizer tokenizes the input sentences, and a CCG parser converts them into CCG trees.
CCG is a grammar formalism that assigns a syntactic category to each grammatical expression.
The set of syntactic categories is defined recursively as follows: (i) atomic categories: $\NP$ (noun phrase), $S$ (sentence), etc., (ii) functional categories: $X / Y$, $X \bs Y$ (where $X$ and $Y$ are syntactic categories).
Both $X / Y$ and $X \bs Y$ take the category $Y$ as an argument and return the category $X$.
``$/$'' and ``$\bs$'' indicate that the argument is taken from the right and left, respectively.

CCG parsers generally use CCGbank~\citep{hockenmaier2007ccgbank} or its modified versions for training, which are not necessarily compatible with comparatives.
Thus, the output CCG trees are not always the ones we expect at this point.
To deal with this issue, we modify the CCG trees if necessary.
Another possible way to modify CCG trees is to revise the CCG parser itself.
However, this method is costly because it requires re-training or fine-tuning the CCG parser.
Thus, we leave this approach for future work.

\subsection{Semantic Parsing}
\label{subsubsec:semantic-parsing}
In this step, we assign a semantic representation to each lexical item of the CCG tree based on the semantic templates.
Then, the semantic representation of the whole sentence is composed according to the CCG rules.
To illustrate, we show two rules below.
Some other rules are provided in \Cref{apd:ccg_rules}.
\begin{itemize}
  \item Forward functional application rule\\
  \begin{minipage}[h]{\linewidth}
    \begin{prooftree}
      \AXC{$\catLF{X / Y}{f}$}
      \AXC{$\catLF{Y}{a}$}
      \RL{$>$}
      \BIC{$\catLF{X}{f\,a}$}
    \end{prooftree}
  \end{minipage}

  \item Backward functional application rule \\
  \begin{minipage}[h]{\linewidth}
    \begin{prooftree}
      \AXC{$\catLF{Y}{a}$}
      \AXC{$\catLF{X \bs Y}{f}$}
      \RL{$<$}
      \BIC{$\catLF{X}{f\,a}$}
    \end{prooftree}
  \end{minipage}

\end{itemize}

We set up the semantic templates in order to give semantic representations to the lexical items.
\Cref{tab:semantic_templates} shows the semantic templates for basic comparative expressions.\footnote{
    For expository purposes, the semantic templates listed here are simplified from the original ones, which are more complicated in order to handle various expressions.
}

Let us proceed to some details of the templates in \Cref{tab:semantic_templates}, focusing on the function $N$ that appears in the templates for positive/negative adjectives, which we have newly added to handle comparatives.\footnote{
    $E$ (resp.\ $Q$) represents the surface form of the word (resp.\ the generalized quantifier~\citep{barwise1981generalized}).
}
$N$ has five arguments, the first one $E$ being the base form of the adjective, and the fifth one $x$ being the subject of the adjective.
Turning to the second argument $\lambda d.d$, it is introduced for the differential comparatives.
Consider (\ref{ex:differential_comparatives_positive}) for example.
In the semantic composition process, this argument becomes $\lambda d. (d + 5)$ as a result of the combination of ``5 kg'' and the adjective ``omoi'' (heavy), which leads to the intended semantic representation (\ref{sr:differential_comparatives_positive}).

\begin{exe}
    \ex\begin{xlist}
        \ex\label{ex:differential_comparatives_positive}
            Taro-wa Jiro yori 5 kg omoi. (Taro is 5 kg heavier than Jiro.)
        \ex\label{sr:differential_comparatives_positive} $\forall d. ( \heavy(\jiro, d) \rightarrow \heavy(\taro, d + 5) )$
    \end{xlist}
\end{exe}

\noindent
The third argument, $\lambda d.d$ (or $\lambda d.{-d}$), indicates whether the adjective is positive or negative.
This allows us to distinguish between (\ref{ex:differential_comparatives_positive})  and (\ref{ex:differential_comparatives_negative}), which contain adjectives of the opposite polarity.
For instance, by assuming that ``karui'' (light) is a negative adjective, we can derive the semantic representation (\ref{sr:differential_comparatives_negative}) for (\ref{ex:differential_comparatives_negative}), where the argument $\lambda d. {-d}$ corresponds to $-5$.
\begin{exe}
    \ex\label{ex:differential_comparatives_polarity}
    \begin{xlist}
        \ex\label{ex:differential_comparatives_negative}
            Taro-wa Jiro yori 5 kg karui. (Taro is 5 kg lighter than Jiro.)
        \ex\label{sr:differential_comparatives_negative} $\forall d. \left( \light(\jiro, d) \rightarrow \light(\taro, d - 5) \right)$
    \end{xlist}
\end{exe}
Similarly, the fourth argument, $\lambda t.t$ (or $\lambda t.\lnot t$), makes a distinction about the polarity of the adjectives in comparatives with measure phrases.
Taking (\ref{ex:measure_phrase_positive}) and (\ref{ex:measure_phrase_negative}) for example, the arguments $\lambda t.t$ and $\lambda t. \lnot t$ correspond to $d > 70$ and $\lnot(d > 70)$ in the semantic representations, respectively.
\begin{exe}
    \ex\label{ex:measure_phrase_positive}
    \begin{xlist}
        \ex
            Taro-wa 70 kg yori omoi. (Taro is heavier than 70 kg.)
        \ex $\exists d. \left( \heavy(\taro, d) \land d > 70 \right)$
    \end{xlist}
    \ex\label{ex:measure_phrase_negative}
    \begin{xlist}
        \ex
            Taro-wa 70 kg yori karui. (Taro is lighter than 70 kg.)
        \ex $\exists d. \left( \light(\taro, d) \land \lnot(d > 70) \right)$
    \end{xlist}
\end{exe}

\subsection{Theorem Proving}
\label{subsec:theorem-proving}
In this step, we input the logical formulas of the premises and hypothesis obtained in the previous step into an automated theorem prover and judge their entailment relation.

\paragraph{Axioms}
In order to prove entailment relations, we introduce some axioms.
To illustrate, we describe one of the axioms, (CP), which is shown below (here, $\bA$ is an adjective).
It corresponds to a basic axiom in degree semantics called Consistency Postulate~\citep{klein1980semantics}.
\begin{description}
    \item[(CP)]
    $\begin{aligned}[t]
    &\forall x\ y. \left( \left( \exists d. \left( \bA(x, d) \land \lnot \bA(y, d) \right) \right) \right.\\
    &{}\rightarrow \forall d. \left.\left( \bA(y, d) \rightarrow \bA(x, d) \right) \right)
    \end{aligned}$
\end{description}
Intuitively, this axiom requires that $\bA$ be a predicate such that if the degree of $x$ is greater than the degree of $y$, then the degree of $x$ is greater than or equal to the degree of $y$.
Using this axiom, we can make inferences such as (\ref{comp-example}).
We give the details of the proof in \Cref{apd:axioms}, where we also explain other axioms.

\paragraph{Implementation}
First, we choose some axioms based on the adjectives in the input sentences and add them as premises.
Then, we input the logical formulas of the premises and hypothesis into the automated theorem prover.
Given the premises and axioms $P_1,\ldots P_n$ and the hypothesis $H$, the system output is \textit{yes} (entailment) when $P_1 \land \dots \land P_n \rightarrow H$ is proven, \textit{no} (contradiction) when $P_1 \land \dots \land P_n \rightarrow \lnot H$ is proven, and \textit{unknown} (neutral) when neither is proven.

\section{Challenges in Handling Japanese Comparatives}
\label{sec:semantic-composition}
In this section, we explain some linguistic phenomena specific to Japanese comparatives and how we treat them in this study.

\subsection{Absence of Overt Comparative Morphemes}
\label{subsec:no_overt}
English has overt comparative morphemes, such as \textit{more} and \textit{-er}.
On the other hand, Japanese has no such morphemes.
The examples (\ref{ex:no_overt_1}) and (\ref{ex:no_overt_2}) illustrate that the adjective ``omoi'' has the same surface form whether it is used for comparison or not.
\begin{exe}
    \ex\label{ex:no_overt}
    \begin{xlist}
        \ex\label{ex:no_overt_1} \gll Taro-wa Jiro yori \underline{omoi}.\\
                 Taro-\Top{} Jiro than heavy\\
                 \glt ``Taro is heavier than Jiro.''
        \ex\label{ex:no_overt_2} \gll Taro-wa \underline{omoi}.\\
                 Taro-\Top{} heavy\\
                 \glt ``Taro is heavy.''
    \end{xlist}
\end{exe}
Although it is possible to give different semantic representations to ``omoi'' in both sentences, we assign the same semantic representation to simplify the semantic parsing process.
Accordingly, we introduce an unpronounced symbol (empty category) to distinguish the semantic representations of the two sentences.
Specifically, when there is no comparative expression such as ``\dots yori'' and ``\dots izyoo-ni,'' we insert an empty category \textit{cmp} of category $S / S$ instead.
We introduce the aforementioned comparison criterion $\theta$ by assigning the following semantic representation (\ref{ex:empty_category_semantics}) to this empty category.
\begin{exe}
    \ex\label{ex:empty_category_semantics}
    $\lambda S. S( \lambda A\ x.A(x, \theta))$
\end{exe}
This inserted operator also plays a role of matching the types of the semantic representations of ``Jiro yori omoi'' and ``\textit{cmp} omoi'' (\Cref{fig:sem-comp-no-overt-1,fig:sem-comp-no-overt-2}).

\begin{figure*}
    \begin{prooftree}
        \small
        \AXC{Jiro}\UIC{$\catLFbr{\NP}{\lambda P. P(\jiro)}$}
        \AXC{yori (than)}\UIC{$\catLFbr{(S/S) \bs \NP}{\lambda Q\,S. S(\lambda A\,x. Q(\lambda y. \exists d. (A(x, d) \land \lnot A(y, d))))}$}
        \RL{$<$}
        \BIC{$\catLFbr{S / S}{\lambda S. S(\lambda A\,x. \exists d. (A(x, d) \land \lnot A(\jiro, d)))}$}
        \AXC{omoi (heavy)}\UIC{$\catLFbr{S \bs \NP}{\lambda Q\,N. Q(\lambda x.N(\heavy, x))}$}
        \RL{$\cfcf$}
        \BIC{$\catLFbr{S \bs \NP}{\lambda Q. Q(\lambda x. \exists d. (\heavy(x, d) \land \lnot \heavy(\jiro, d)))}$}
    \end{prooftree}
    \vspace{-0.2cm}
    \caption{A part of semantic composition of (\ref{ex:no_overt_1})}
    \label{fig:sem-comp-no-overt-1}

    \vspace{0.5cm}
    \begin{prooftree}
        \AXC{\textit{cmp}}\UIC{$\catLFbr{S / S}{\lambda S. S(\lambda A\,x. A(x, \theta))}$}
        \AXC{omoi (heavy)}\UIC{$\catLFbr{S \bs \NP}{\lambda Q\,N. Q(\lambda x. N(\heavy, x))}$}
        \RL{$\cfcf$}
        \BIC{$\catLFbr{S \bs \NP}{\lambda Q. Q(\lambda x. \heavy(x, \theta))}$}
    \end{prooftree}
    \vspace{-0.2cm}
    \caption{A part of semantic composition of (\ref{ex:no_overt_2})}
    \label{fig:sem-comp-no-overt-2}
\end{figure*}

\subsection{Equatives}
\label{subsec:equatives}
English equative sentences such as (\ref{ex:english_equative}) are interpreted as indicating ``... is at least as heavy as ...~.''
Thus,~\citet{Haruta_Mineshima_Bekki_2022} represented (\ref{ex:english_equative}) as (\ref{ex:english_equative_logic}).
\begin{exe}
    \ex\begin{xlist}
        \ex\label{ex:english_equative} John is as heavy as Bob.
        \ex\label{ex:english_equative_logic}
    $\forall d. \left( \heavy(\bob, d) \rightarrow \heavy(\john, d) \right)$
    \end{xlist}
\end{exe}
On the other hand, Japanese equatives merely express that the degrees are close to each other. For instance, (\ref{jei-1}) can be true even when Taro's weight is slightly less than Jiro's. 
\begin{exe}
    \ex\label{ex:japanese_equative_inference}
    \begin{xlist}
        \ex\label{jei-1}
            \gll Taro-wa Jiro to onaji kurai-no omosa-da.\\
                 Taro-\Top{} Jiro {} same as-\Gen{} weight-\Cop{}\\
            \glt ``Taro is as heavy as Jiro.''
        \ex\label{jei-2}
            \gll Jiro-wa omoi.\\
                 Jiro-\Top{} heavy\\
            \glt ``Jiro is heavy.''
        \ex\label{jei-3}
            \gll Taro-wa omoi. (entailment)\\
                 Taro-\Top{} heavy\\
            \glt ``Taro is heavy.''
    \end{xlist}
\end{exe}
To handle the meaning of equatives, we propose the following representation (\ref{ex:japanese_equative_logic}) for (\ref{jei-1}).
This intuitively indicates that the difference in weight between Taro and Jiro is less than the constant $\delta$.
\begin{exe}
    \ex\label{ex:japanese_equative_logic}
    $\begin{aligned}[t]
        &\forall d_1 d_2. \left( \left( \lnot\left( \heavy(\taro, d_1) \right.\right.\right. \\
        &{}\leftrightarrow \left.\heavy(\jiro, d_1) \right) \\
        &{}\land \left.\lnot\left( \heavy(\taro, d_2) \leftrightarrow \heavy(\jiro, d_2) \right) \right) \\
        &{}\rightarrow \left.| d_1 - d_2 | < \delta \right)
    \end{aligned}$
\end{exe}
We also introduce the following axiom (\ref{ex:axiom_delta}), which prescribes the relation between $\theta$ and $\delta$.
Intuitively, this axiom indicates that $\delta$ is so small that the truth value of the predicate $\heavy$ does not change within the range of $\delta$ from $\theta$.
\begin{exe}
    \ex\label{ex:axiom_delta}
    $\forall x. \left( \heavy(x, \theta - \delta) \leftrightarrow \heavy(x, \theta + \delta) \right)$
\end{exe}
We can make inferences such as (\ref{ex:japanese_equative_inference}) using this axiom together with (UP) and (DOWN) (see~\Cref{apd:axioms} for details).

\subsection{Clausal Comparatives}
\label{subsec:clausal_comparatives}
\begin{table*}[h]
  \centering
  \begin{tabular}{lll}
    \hline
    Category & Word & Semantic Template\\
    \hline \hline
    \begin{tabular}[c]{@{}l@{}}
      $\left( \left( \NP / \NP \right) / \left( \NP / \NP \right) \right) \bs \left( S \bs \NP \right)$
    \end{tabular}
    &
    \begin{tabular}[c]{@{}l@{}}
      yori \\
      (\ref{ex:clausal_comparatives_1})
    \end{tabular}
    &
    \begin{tabular}[c]{@{}l@{}}
      $\lambda E\,V\,M. V(\lambda G.\exists d. (M(\lambda A\,x. A(x, d)) $\\
      ${}\land \lnot M(\lambda A\,x. (A(x, d) \land G(x)))))$
    \end{tabular}
    \\
    \hline
    \begin{tabular}[c]{@{}l@{}}
      $\left( \left( \NP / \NP \right) / \left( \NP / \NP \right) \right) \bs \NP$
    \end{tabular}
    &
    \begin{tabular}[c]{@{}l@{}}
      yori \\
      (\ref{ex:clausal_comparatives_2})
    \end{tabular}
    &
    \begin{tabular}[c]{@{}l@{}}
      $\lambda E\,Q\,M. \exists d. (M(\lambda A\,x. A(x, d))$\\
      ${}\land \lnot M(\lambda A\,x.(A(x, d) \land Q(\lambda y. (x = y)))))$
    \end{tabular}
    \\
    \hline
    \begin{tabular}[c]{@{}l@{}}
      $\left( \left( \NP / \NP \right) / \left( \NP / \NP \right) \right) \bs \NP$
    \end{tabular}
    &
    \begin{tabular}[c]{@{}l@{}}
      yori \\
      (\ref{ex:clausal_comparatives_3})
    \end{tabular}
    &
    \begin{tabular}[c]{@{}l@{}}
      $\lambda E\,Q\,M\,F\,x. Q(\lambda y.$\\
      $(\exists d. M(\lambda A\,z. (A(z, d) \land F(x, z))) $\\
      ${}\land \lnot M(\lambda A\,z. (A(z, d) \land F(y, z)))))$
    \end{tabular}
    \\
    \hline
  \end{tabular}
  \caption{Semantic templates for clausal comparatives}
  \label{tab:clausal_semantic_templates}
\end{table*}

Clausal comparatives are comparatives with subordinate clauses.
(\ref{ex:clausal_comparatives_1}) is an example of a clausal comparative.
We also deal with related sentences such as (\ref{ex:clausal_comparatives_2}) and (\ref{ex:clausal_comparatives_3}).
\begin{exe}
    \ex\label{ex:clausal_comparatives}
    \begin{xlist}
        \ex\label{ex:clausal_comparatives_1}
        \gll Taro-wa Hanako-ga katta yori takai hon-o katta.\\
                 Taro-\Top{} Hanako-\Nom{} bought than expensive book-\Acc{} bought\\
        \glt ``Taro bought a more expensive book than Hanako bought.''
        \ex\label{ex:clausal_comparatives_2}
        \gll Taro-wa Hanako-ga katta no yori takai hon-o katta.\\
                 Taro-\Top{} Hanako-\Nom{} bought \No{} than expensive book-\Acc{} bought\\
        \glt ``Taro bought a more expensive book than what Hanako bought.''
        \ex\label{ex:clausal_comparatives_3}
        \gll Taro-wa Hanako yori takai hon-o katta.\\
                 Taro-\Top{} Hanako than expensive book-\Acc{} bought\\
        \glt ``Taro bought a more expensive book than Hanako.''
    \end{xlist}
\end{exe}
We assign the same semantic representation (\ref{ex:clausal_comparatives_semantics}) to the three sentences in (\ref{ex:clausal_comparatives}).
\begin{exe}
    \ex\label{ex:clausal_comparatives_semantics}
    $\begin{aligned}[t]
        &\exists d. \left( \exists x. \left( \book(x) \land \expensive(x, d) \right.\right.\\
        &{}\land \exists e. \left( \bought(e) \land \left( \nom(e) = \taro \right)\right. \\
        &{}\land \left.\left.\left( \acc(e) = x \right) \right) \right)\\
        &{}\land \lnot \exists x. \left( \book(x) \land \expensive(x, d) \right.\\
        &{}\land \exists e. \left( \bought(e) \land \left( \nom(e) = \hanako \right)\right. \\
        &\land \left.\left.\left.\left( \acc(e) = x \right) \right) \right) \right)
    \end{aligned}$
\end{exe}
In order to obtain this semantic representation, we assign different semantic representations to ``yori'' in each sentence, which are listed in Table \ref{tab:clausal_semantic_templates}.
Note that the template in the second row includes $\lambda y. (x = y)$, which is necessary to consider the fact that the pronominal ``no'' is identified with ``hon'' (book) in (\ref{ex:clausal_comparatives_2}).

\subsection{Presupposition}
\label{subsec:presupposition}
Some Japanese comparative expressions have a special semantic content called a presupposition \citep{kubota2012presuppositional,hayashishita2007izyoo}.
A presupposition is a type of meaning not affected by entailment-canceling operators such as negation and modals (cf. \citet{potts2015presupposition}).
The predicate ``know'' is an example of a presupposition trigger (i.e., an expression or a construction causing presuppositions).
In (\ref{ex:presupposition_know}), the presupposition is that Bob ran.
This can be confirmed by the fact that the negated sentence (\ref{ex:presupposition_know_negation}) also implies that Bob ran.

\begin{exe}
    \ex\begin{xlist}
        \ex\label{ex:presupposition_know}
    John knows that Bob ran.
        \ex\label{ex:presupposition_know_negation}
    John does not know that Bob ran.
    \end{xlist}
\end{exe}

\noindent
We list some Japanese comparative sentences with a presupposition in (\ref{ex:jp_presupposition}), where the trigger is underlined.
Here, the presupposition is that the comparative standard has the property expressed by the predicate.
That is, the three sentences in (\ref{ex:jp_presupposition}) all presuppose that Jiro is heavy.
\begin{exe}
    \ex\label{ex:jp_presupposition}
    \begin{xlist}
        \ex\label{jp_presupposition_1}
        \gll Taro-wa Jiro \underline{izyoo-ni} omoi.\\
                 Taro-\Top{} Jiro than heavy\\
        \glt ``Taro is heavier than Jiro.''
        \ex\label{jp_presupposition_2}
        \gll Taro-wa Jiro \underline{to onaji\, kurai} omoi.\\
                 Taro-\Top{} Jiro {as same as} heavy\\
        \glt ``Taro is as heavy as Jiro.''
        \ex\label{jp_presupposition_3}
        \gll Taro-wa Jiro \underline{hodo} omoku \underline{nai}.\\
                 Taro-\Top{} Jiro hodo heavy not\\
        \glt ``Taro is not as heavy as Jiro.''
    \end{xlist}
\end{exe}

In formally analyzing presuppositions, it is not adequate to simply conjoin the presupposition with other parts of the sentence.
For example, suppose we represent the meaning of (\ref{ex:presupposition_know}) as a conjunction of the semantic representations of ``John knows that Bob ran'' and ``Bob ran,'' as shown below.

\begin{exe}
    \ex\label{ex:presupposition_know_semantics}
    $\know(\john, \ran(\bob)) \land \ran(\bob)$
\end{exe}

\noindent
The negation of this formula, which is shown in (\ref{ex:presupposition_know_negation_semantics}), 
does not entail $\ran(\bob)$, failing to capture the fact that the presupposition is not subject to the negation (cf. (\ref{ex:presupposition_know_negation})).

\begin{exe}
    \ex\label{ex:presupposition_know_negation_semantics}
    $\begin{aligned}[t]
    &\lnot \left( \know(\john, \ran(\bob)) \land \ran(\bob) \right)\\
    &{}\Leftrightarrow \lnot \know(\john, \ran(\bob)) \lor \lnot \ran(\bob)\end{aligned}$
\end{exe}

In order to correctly handle presuppositions, we use a framework called \textit{multidimensional semantics} \citep{karttunen1979conventional}.
In this framework, the semantic representation of an entire sentence is represented by a pair of semantic representations.
The first element is for the central content conveyed by the sentence (the \textit{at-issue} content), and the second one is for the presupposition.
For example, the semantic representation of the sentence (\ref{jp_presupposition_1}) is shown in (\ref{jp_presupposition_1_semantics}).
\begin{exe}
    \ex\label{jp_presupposition_1_semantics}
    $\begin{aligned}[t]
        &\langle \exists d. \left( \heavy(\taro, d) \land \lnot \heavy(\jiro, d) \right),\\
        &\heavy(\jiro, \theta) \rangle
    \end{aligned}$
\end{exe}
When the sentence is negated, we only negate the semantic representation of the at-issue content in the semantic composition, and the semantic representation of the entire sentence is (\ref{jp_presupposition_1_negation_semantics}).
\begin{exe}
    \ex\label{jp_presupposition_1_negation_semantics}
    $\begin{aligned}[t]
        &\langle \lnot \exists d. \left( \heavy(\taro, d) \land \lnot \heavy(\jiro, d) \right),\\
        &\heavy(\jiro, \theta) \rangle
    \end{aligned}$
\end{exe}

\noindent
In the theorem proving step, we conjoin the semantic representations for the at-issue content and for the presupposition with $\land$.

\section{Experiment}
\subsection{Settings}
\label{sec:settings}
In this section, we describe the implementation settings of the proposed system.

\paragraph{Syntactic Parsing}
We use a Japanese tokenizer Janome.\footnote{\url{https://github.com/mocobeta/janome}}
As a CCG parser, we use depccg~\citep{yoshikawa:2017acl}, the best-performing model provided for Japanese.
We use Tsurgeon~\citep{levy2006tregex} to modify CCG parsing trees and insert empty categories.
Our modification processes are as follows:
\begin{itemize}
  \item We add rules to merge some multiword expressions. For instance, ``izyoo ni'' is converted to ``izyoo-ni,'' ``yori mo'' to ``yori-mo,'' and ``to onaji kurai no'' to ``to-onaji-kurai-no.''
  \item We insert the empty category \textit{cmp} (cf.~\Cref{subsec:no_overt}).
  \item We add a new syntactic feature to ``yori'' in phrasal comparatives related to clausal comparatives\footnote{For example, ``Taro-wa Hanako yori takai hon-o katta. (Taro bought a more expensive book than Hanako.)''} in order to distinguish it from ``yori'' in ordinary phrasal comparatives.
  \item We add a new syntactic feature to ``yori'' in comparatives with a measure phrase in order to distinguish it from ``yori'' in ordinary phrasal comparatives.
\end{itemize}
In total, we make 60 entries in the Tsurgeon script for these processes.

\paragraph{Semantic Parsing}
For semantic composition, we use ccg2lambda \citep{martinez2016ccg2lambda}, which supports Japanese as well as English.
It uses $\lambda$-calculus to derive semantic representations.
We extend the semantic templates to introduce the semantic representations based on degree semantics.
We create two templates, one with multidimensional semantics and one without.
The total number of lexical entries in each semantic template file is 222.
We newly add 58 entries for words related to comparatives.

\paragraph{Theorem Proving}
We use Vampire 4.9~\citep{kovacs2013first}, a resolution-based automated theorem prover, for theorem proving.
Vampire uses the Thousand of Problems for Theorem Provers (TPTP,~\citealt{sutcliffe2017tptp}) format to describe logical formulas.
For this reason, we convert the output of ccg2lambda into first-order predicate logic formulas in the TPTP format.
At this point, we add the axioms described in Section \ref{subsec:theorem-proving}.
In this step, we use the CASC mode, the fastest mode in Vampire.
We try to prove $P_1 \land P_2 \land \dots \land P_n \rightarrow H$ and $P_1 \land P_2 \land \dots \land P_n \rightarrow \lnot H$ for up to 20 seconds each to determine the system output.

\subsection{Dataset}
\label{sec:dataset}
We use the comparatives section of the JSeM dataset~\cite{10.1007/978-3-319-50953-2_5} for evaluation of our inference system.
This NLI dataset contains Japanese counterparts of the FraCaS test suite~\cite{cooper1996using}.
It also contains newly added problems that involve phenomena FraCaS does not address or phenomena unique to Japanese.

In this study, we do not address tense and aspect, so we eliminated problems involving them.
We do not address modality as well.
With regard to modality, JSeM only has problems involving the property that modals do not affect the presupposition.
Thus, we replaced modals with negation on these problems.
As a result, the number of problems in the dataset is 71.
The distribution of the gold answer labels is (\textit{yes}/\textit{no}/\textit{unknown}) = (42/8/21).
Table \ref{tab:jsem-example} shows some problems in the dataset.
\begin{table}[h]
  \centering
  \begin{tabular}{lp{6.5cm}}
    \hline
    \multicolumn{2}{l}{jsem-569, Gold answer: yes}\\
    \hline \hline
    P1 &
    \begin{tabular}{p{6.2cm}}
      PC-6082-wa ITEL-XZ yori hayai.\\
      (PC-6082 is faster than ITEL-XZ.)
    \end{tabular}
    \\
    P2 &
    \begin{tabular}{l}
      ITEL-XZ-wa hayai.\\
      (ITEl-XZ is fast.)
    \end{tabular}
    \\
    H &
    \begin{tabular}{l}
      PC-6082-wa hayai.\\
      (PC-6082 is fast.)
    \end{tabular}
    \\
    \hline
    \multicolumn{2}{l}{jsem-576, Gold answer: no}\\
    \hline \hline
    P1 &
    \begin{tabular}{p{6.2cm}}
      PC-6082-wa ITEL-XZ to onaji kurai-no hayasa-da.\\
      (PC-6082 is as fast as ITEL-XZ.)
    \end{tabular}
    \\
    P2 &
    \begin{tabular}{l}
      PC-6082-wa osoi.\\
      (PC-6082 is slow.)
    \end{tabular}
    \\
    H &
    \begin{tabular}{l}
      ITEL-XZ-wa hayai.\\
      (ITEL-XZ is fast.)
    \end{tabular}
    \\
    \hline
  \end{tabular}
  \caption{Examples of the problems in JSeM. P and H stand for ``premise'' and ``hypothesis,'' respectively.}
  \label{tab:jsem-example}
\end{table}

\subsection{Evaluation Method}
\label{sec:evaluation-method}
We use accuracy as an evaluation metric.
When an error occurs in the proposed system, we treat it as an incorrect answer.
As a baseline, we adopt GPT-4o and Swallow 8B\footnote{tokyotech-llm/Llama-3.1-Swallow-8B-v0.1}/70B\footnote{tokyotech-llm/Llama-3.1-Swallow-70B-v0.1} (S-8B/S-70B), the latter being competitive open Japanese LLMs.
We conduct experiments using six different prompts for these models and calculate the accuracy as the average across these prompts.\footnote{
    The model inferences were conducted in May 2025.
}
The details of the prompts are shown in \Cref{apd:prompts}.

\section{Results and Discussion}
\label{chap:discussion}

\subsection{Results}
\label{sec:results}
Table \ref{tab:results} shows the accuracy on the JSeM dataset.
The table shows that our system outperformed all baseline models in terms of accuracy.
The detailed results are shown in \Cref{apd:results}.
\begin{table}[h]
  \centering
  \begin{tabular}{ccccc}
    \hline
    Majority & GPT-4o & S-8B & S-70B & Ours \\
    \hline
    .592 & .774 & .549 & .712 & \textbf{.845} \\
    \hline
  \end{tabular}
  \caption{
        Accuracy on the JSeM dataset.
        ``Majority'' indicates the accuracy achieved when ``yes,'' the most common label in the dataset, is answered for all the problems.
    }
  \label{tab:results}
\end{table}

\begin{table}[h]
  \centering
  \begin{tabular}{ll}
    \hline
    \multicolumn{2}{l}{jsem-570, Gold answer: unknown}\\
    \multicolumn{2}{l}{GPT-4o: yes, Ours: unknown}\\
    \hline \hline
    P &
    \begin{tabular}{l}
      PC-6082-wa ITEL-XZ yori hayai.\\
      (PC-6082 is faster than ITEL-XZ.)
    \end{tabular}
    \\
    H &
    \begin{tabular}{l}
      PC-6082-wa hayai.\\
      (PC-6082 is fast.)
    \end{tabular}
    \\
    \hline
    \multicolumn{2}{l}{jsem-620, Gold answer: yes}\\
    \multicolumn{2}{l}{GPT-4o: unknown, Ours: yes}\\
    \hline \hline
    P &
    \begin{tabular}{l}
      Taro-wa Hanako izyoo-ni hayaoki-da.\\
      (Taro is an earlier riser than Hanako.)
    \end{tabular}
    \\
    H &
    \begin{tabular}{l}
      Hanako-wa hayaoki-da.\\
      (Hanako is an early riser.)
    \end{tabular}
    \\
    \hline
  \end{tabular}
  \caption{Examples of problems that GPT-4o did not answer correctly but ours did}
  \label{tab:comparison-gpt-ours}
\end{table}

Table \ref{tab:comparison-gpt-ours} shows some examples of problems that our system could predict correct answers while GPT-4o could not.
GPT-4o incorrectly answered some of the relatively simple problems, such as jsem-570.
The possible reason is that GPT-4o inferred ``X is fast'' from ``X is faster.''

\noindent
Notably, GPT-4o failed to answer correctly some problems with presupposition triggers, such as jsem-620.
In order to perform this inference, it is necessary to infer the presupposition that Hanako is an early riser from the premise.
GPT-4o was rarely able to solve such problems.
On the other hand, our proposed system correctly predicted the entailment relation, thanks to multidimensional semantics.

\subsection{Error Analysis}
\label{sec:error-analysis}
\Cref{tab:error-examples} shows two cases where our system failed to obtain correct semantic representations, but GPT-4o gave correct answers.
In jsem-589, we can interpret ``APCOM-no keiyaku'' either as the contracts that APCOM won or as the contracts that ITEL won from APCOM.
To handle this kind of ambiguity, we need to (i) add a new semantic representation of ``yori,'' and (ii) implement a system for distinguishing between the two interpretations based on syntactic information.

Jsem-606 is another case where our system failed to make a correct prediction.
The verb ``magaru'' (bend) behaves like an adjective when combined with ``te-i-ru.''
However, our system treats the resultant predicate ``magatte-i-ru'' as a verb, so its semantic type does not match the one required for the argument of ``yori,'' causing an error in semantic parsing.
To handle this error, we need to give an exceptional semantic representation to ``te-i-ru'' when it forms an adjective-like predicate with certain verbs like ``magaru.''

\begin{table}[h]
  \centering
  \begin{tabular}{lp{6.5cm}}
    \hline
    \multicolumn{2}{l}{jsem-589, Gold answer: yes}\\
    \multicolumn{2}{l}{GPT-4o: yes, Ours: error}\\
    \hline \hline
    P &
    \begin{tabular}{p{6.2cm}}
      ITEL-wa APCOM-no keiyaku yori ooku-no
      chuumon-o kakutoku-sita.\\
      (ITEL won more orders than the APCOM contract.)
    \end{tabular}
    \\
    H &
    \begin{tabular}{p{6.2cm}}
      ITEL-wa APCOM-no chuumon-o kakutoku-shita.\\
      (ITEL won the APCOM contract.)
    \end{tabular}
    \\
    \hline
    \multicolumn{2}{l}{jsem-606, Gold answer: yes}\\
    \multicolumn{2}{l}{GPT-4o: yes, Ours: error}\\
    \hline \hline
    P &
    \begin{tabular}{l}
      Kono boo-wa ano boo yori magatte-i-ru.\\
      (This stick is more bent than that one.)
    \end{tabular}
    \\
    H &
    \begin{tabular}{l}
      Kono boo-wa magatte-i-ru.\\
      (This stick is bent.)
    \end{tabular}
    \\
    \hline
  \end{tabular}
  \caption{Examples of problems our system answered incorrectly}
  \label{tab:error-examples}
\end{table}

\section{Conclusion}
\label{chap:conclusion-and-future-work}

In this study, we have proposed ccg-jcomp, a logical inference system for Japanese comparatives based on CCG, degree semantics, and some analyses of phenomena unique to Japanese comparatives.
In our experiments with the Japanese NLI dataset that involves comparatives, we demonstrated that our proposed system achieved higher accuracy than several LLMs.

In future work, we are considering handling the ambiguity of certain sentences and the behavior of the adjective-like verbs discussed in \Cref{sec:error-analysis}.
Additionally, it would be desirable to address adverbial comparatives, which are not covered in JSeM.

\section*{Limitations}
\paragraph{Few-shot Learning}
In this study, we did not compare methods using few-shot learning as a baseline.
It may improve the performance of the baseline models.
For example, the LLMs may correctly answer jsem-620 in \Cref{sec:results} by looking at some example inferences with a presupposition and learning the inference patterns.
However, we do not have a sufficient number of problems involving Japanese comparatives to carry out and evaluate few-shot learning.
Therefore, we conducted all experiments in a zero-shot setting for all models.

\paragraph{Scalability}
In addition to comparatives, JSeM has sections on other linguistic phenomena, such as anaphora.
However, since our proposed system focuses only on Japanese comparatives, it cannot be used as is to handle these phenomena.
To address them, we need to introduce the mechanism employed by some specific frameworks (e.g., \textit{dynamic semantics}~\citep{GroenendijkStokhof1991} for anaphora) in a manner consistent with degree semantics, which is not trivial.
Hence, we leave for future work the development of a unified system that can handle these phenomena together with comparatives.

\section*{Acknowledgments}
We would like to thank the anonymous reviewers for their valuable comments, which helped us improve this paper.
This work was supported by the Institute for AI and Beyond of the University of Tokyo and JSPS KAKENHI Grant Number JP24H00809, Japan.

\bibliography{latex/myref}

\begin{thebibliography}{29}
\providecommand{\natexlab}[1]{#1}

\bibitem[{Abzianidze(2015)}]{abzianidze-2015-tableau}
Lasha Abzianidze. 2015.
\newblock \href {https://doi.org/10.18653/v1/D15-1296} {A tableau prover for
  natural logic and language}.
\newblock In \emph{Proceedings of the 2015 Conference on Empirical Methods in
  Natural Language Processing}, pages 2492--2502, Lisbon, Portugal. Association
  for Computational Linguistics.

\bibitem[{Barwise and Cooper(1981)}]{barwise1981generalized}
Jon Barwise and Robin Cooper. 1981.
\newblock Generalized quantifiers and natural language.
\newblock In \emph{Philosophy, language, and artificial intelligence: Resources
  for processing natural language}, pages 241--301. Springer.

\bibitem[{Bernardy and
  Chatzikyriakidis(2017)}]{bernardy-chatzikyriakidis-2017-type}
Jean-Philippe Bernardy and Stergios Chatzikyriakidis. 2017.
\newblock \href {https://aclanthology.org/W17-6801/} {A type-theoretical system
  for the {F}ra{C}a{S} test suite: Grammatical framework meets coq}.
\newblock In \emph{Proceedings of the 12th International Conference on
  Computational Semantics ({IWCS}) {---} Long papers}.

\bibitem[{Bernardy and
  Chatzikyriakidis(2021)}]{bernardy-chatzikyriakidis-2021-applied}
Jean-Philippe Bernardy and Stergios Chatzikyriakidis. 2021.
\newblock \href {https://aclanthology.org/2021.iwcs-1.2/} {Applied temporal
  analysis: A complete run of the {F}ra{C}a{S} test suite}.
\newblock In \emph{Proceedings of the 14th International Conference on
  Computational Semantics (IWCS)}, pages 11--20, Groningen, The Netherlands
  (online). Association for Computational Linguistics.

\bibitem[{Bowman et~al.(2015)Bowman, Angeli, Potts, and
  Manning}]{bowman-etal-2015-large}
Samuel~R. Bowman, Gabor Angeli, Christopher Potts, and Christopher~D. Manning.
  2015.
\newblock \href {https://doi.org/10.18653/v1/D15-1075} {A large annotated
  corpus for learning natural language inference}.
\newblock In \emph{Proceedings of the 2015 Conference on Empirical Methods in
  Natural Language Processing}, pages 632--642, Lisbon, Portugal. Association
  for Computational Linguistics.

\bibitem[{Cooper et~al.(1996)Cooper, Crouch, Van~Eijck, Fox, Van~Genabith,
  Jaspars, Kamp, Milward, Pinkal, Poesio et~al.}]{cooper1996using}
Robin Cooper, Dick Crouch, Jan Van~Eijck, Chris Fox, Johan Van~Genabith, Jan
  Jaspars, Hans Kamp, David Milward, Manfred Pinkal, Massimo Poesio, et~al.
  1996.
\newblock Using the framework.
\newblock Technical report, Technical Report LRE 62-051 D-16, The FraCaS
  Consortium.

\bibitem[{Cresswell(1976)}]{cresswell1976semantics}
Max~J Cresswell. 1976.
\newblock The semantics of degree.
\newblock In \emph{Montague grammar}, pages 261--292. Elsevier.

\bibitem[{Groenendijk and Stokhof(1991)}]{GroenendijkStokhof1991}
Jeroen Groenendijk and Martin Stokhof. 1991.
\newblock \href {https://doi.org/10.1007/BF00628304} {Dynamic predicate logic}.
\newblock \emph{Linguistics and Philosophy}, 14(1):39--100.

\bibitem[{Haruta et~al.(2022)Haruta, Mineshima, and
  Bekki}]{Haruta_Mineshima_Bekki_2022}
Izumi Haruta, Koji Mineshima, and Daisuke Bekki. 2022.
\newblock \href {https://doi.org/10.15398/jlm.v10i1.294} {Implementing natural
  language inference for comparatives}.
\newblock \emph{Journal of Language Modelling}, 10(1):139--191.

\bibitem[{Hayashishita(2007)}]{hayashishita2007izyoo}
J-R Hayashishita. 2007.
\newblock Izyoo (ni)-and gurai-comparatives: Comparisons of deviation in
  japanese.
\newblock \emph{GENGO KENKYU (Journal of the Linguistic Society of Japan)},
  132:77--109.

\bibitem[{Hockenmaier and Steedman(2007)}]{hockenmaier2007ccgbank}
Julia Hockenmaier and Mark Steedman. 2007.
\newblock Ccgbank: a corpus of ccg derivations and dependency structures
  extracted from the penn treebank.
\newblock \emph{Computational Linguistics}, 33(3):355--396.

\bibitem[{Hu et~al.(2020)Hu, Chen, Richardson, Mukherjee, Moss, and
  Kuebler}]{hu-etal-2020-monalog}
Hai Hu, Qi~Chen, Kyle Richardson, Atreyee Mukherjee, Lawrence~S. Moss, and
  Sandra Kuebler. 2020.
\newblock \href {https://aclanthology.org/2020.scil-1.40/} {{M}ona{L}og: a
  lightweight system for natural language inference based on monotonicity}.
\newblock In \emph{Proceedings of the Society for Computation in Linguistics
  2020}, pages 334--344, New York, New York. Association for Computational
  Linguistics.

\bibitem[{Karttunen and Peters(1979)}]{karttunen1979conventional}
Lauri Karttunen and Stanley Peters. 1979.
\newblock Conventional lmplicature.
\newblock In \emph{Presupposition}, pages 1--56. Brill.

\bibitem[{Kawazoe et~al.(2017)Kawazoe, Tanaka, Mineshima, and
  Bekki}]{10.1007/978-3-319-50953-2_5}
Ai~Kawazoe, Ribeka Tanaka, Koji Mineshima, and Daisuke Bekki. 2017.
\newblock An inference problem set for evaluating semantic theories and
  semantic processing systems for japanese.
\newblock In \emph{New Frontiers in Artificial Intelligence}, pages 58--65,
  Cham. Springer International Publishing.

\bibitem[{Klein(1980)}]{klein1980semantics}
Ewan Klein. 1980.
\newblock A semantics for positive and comparative adjectives.
\newblock \emph{Linguistics and Philosophy}, 4:1--45.

\bibitem[{Klein(1982)}]{klein1982interpretation}
Ewan Klein. 1982.
\newblock The interpretation of adjectival comparatives1.
\newblock \emph{Journal of Linguistics}, 18(1):113--136.

\bibitem[{Kov{\'a}cs and Voronkov(2013)}]{kovacs2013first}
Laura Kov{\'a}cs and Andrei Voronkov. 2013.
\newblock First-order theorem proving and vampire.
\newblock In \emph{International Conference on Computer Aided Verification},
  pages 1--35. Springer.

\bibitem[{Kubota(2012)}]{kubota2012presuppositional}
Yusuke Kubota. 2012.
\newblock The presuppositional nature of izyoo (-ni) and gurai comparatives: A
  note on hayashishita (2007).
\newblock \emph{GENGO KENKYU (Journal of the Linguistic Society of Japan)},
  141:33--47.

\bibitem[{Levy and Andrew(2006)}]{levy2006tregex}
Roger Levy and Galen Andrew. 2006.
\newblock \href {https://aclanthology.org/L06-1311/} {{Tregex} and {Tsurgeon}:
  tools for querying and manipulating tree data structures}.
\newblock In \emph{Proceedings of the Fifth International Conference on
  Language Resources and Evaluation ({LREC}{'}06)}, Genoa, Italy. European
  Language Resources Association (ELRA).

\bibitem[{Liu et~al.(2023)Liu, Ning, Teng, Liu, Zhou, and
  Zhang}]{liu2023evaluatinglogicalreasoningability}
Hanmeng Liu, Ruoxi Ning, Zhiyang Teng, Jian Liu, Qiji Zhou, and Yue Zhang.
  2023.
\newblock \href {https://arxiv.org/abs/2304.03439} {Evaluating the logical
  reasoning ability of {ChatGPT} and {GPT-4}}.
\newblock \emph{Preprint}, arXiv:2304.03439.

\bibitem[{Mart{\'\i}nez-G{\'o}mez et~al.(2016)Mart{\'\i}nez-G{\'o}mez,
  Mineshima, Miyao, and Bekki}]{martinez2016ccg2lambda}
Pascual Mart{\'\i}nez-G{\'o}mez, Koji Mineshima, Yusuke Miyao, and Daisuke
  Bekki. 2016.
\newblock ccg2lambda: A compositional semantics system.
\newblock In \emph{Proceedings of ACL-2016 System Demonstrations}, pages
  85--90.

\bibitem[{Mineshima et~al.(2015)Mineshima, Mart{\'\i}nez-G{\'o}mez, Miyao, and
  Bekki}]{mineshima2015higher}
Koji Mineshima, Pascual Mart{\'\i}nez-G{\'o}mez, Yusuke Miyao, and Daisuke
  Bekki. 2015.
\newblock Higher-order logical inference with compositional semantics.
\newblock In \emph{Proceedings of the 2015 Conference on Empirical Methods in
  Natural Language Processing}, pages 2055--2061.

\bibitem[{Parmar et~al.(2024)Parmar, Patel, Varshney, Nakamura, Luo, Mashetty,
  Mitra, and Baral}]{parmar-etal-2024-logicbench}
Mihir Parmar, Nisarg Patel, Neeraj Varshney, Mutsumi Nakamura, Man Luo, Santosh
  Mashetty, Arindam Mitra, and Chitta Baral. 2024.
\newblock \href {https://doi.org/10.18653/v1/2024.acl-long.739}
  {{L}ogic{B}ench: Towards systematic evaluation of logical reasoning ability
  of large language models}.
\newblock In \emph{Proceedings of the 62nd Annual Meeting of the Association
  for Computational Linguistics (Volume 1: Long Papers)}, pages 13679--13707,
  Bangkok, Thailand. Association for Computational Linguistics.

\bibitem[{Potts(2015)}]{potts2015presupposition}
Christopher Potts. 2015.
\newblock Presupposition and implicature.
\newblock \emph{The Handbook of Contemporary Semantic Theory}, pages 168--202.

\bibitem[{Seuren(1973)}]{Seuren1973}
Pieter A.~M. Seuren. 1973.
\newblock \href {https://doi.org/10.1007/978-94-010-2503-4_22} {\emph{The
  Comparative}}, pages 528--564.
\newblock Springer Netherlands, Dordrecht.

\bibitem[{She et~al.(2023)She, Potts, Bowman, and Geiger}]{she2023scone}
Jingyuan~S. She, Christopher Potts, Samuel~R. Bowman, and Atticus Geiger. 2023.
\newblock \href {https://doi.org/10.18653/v1/2023.acl-short.154} {{S}co{N}e:
  Benchmarking negation reasoning in language models with fine-tuning and
  in-context learning}.
\newblock In \emph{Proceedings of the 61st Annual Meeting of the Association
  for Computational Linguistics (Volume 2: Short Papers)}, pages 1803--1821,
  Toronto, Canada. Association for Computational Linguistics.

\bibitem[{Steedman(2000)}]{steedman2000syntactic}
Mark Steedman. 2000.
\newblock \emph{The Syntactic Process}.
\newblock MIT press.

\bibitem[{Sutcliffe(2017)}]{sutcliffe2017tptp}
Geoff Sutcliffe. 2017.
\newblock The {{TPTP}} problem library and associated infrastructure: from
  {{CNF}} to {{TH0}}, {{TPTP}} v6. 4.0.
\newblock \emph{Journal of Automated Reasoning}, 59(4):483--502.

\bibitem[{Yoshikawa et~al.(2017)Yoshikawa, Noji, and
  Matsumoto}]{yoshikawa:2017acl}
Masashi Yoshikawa, Hiroshi Noji, and Yuji Matsumoto. 2017.
\newblock \href {https://doi.org/10.18653/v1/P17-1026} {A* {CCG} parsing with a
  supertag and dependency factored model}.
\newblock In \emph{Proceedings of the 55th Annual Meeting of the Association
  for Computational Linguistics (Volume 1: Long Papers)}, pages 277--287.
  Association for Computational Linguistics.

\end{thebibliography}

\appendix

\section{Combinatory Rules of CCG}
\label{apd:ccg_rules}
We show some combinatory rules of CCG below (see~\citet{steedman2000syntactic} for details).
\begin{itemize}
  \item Forward functional application rule\\
  \begin{minipage}[h]{\linewidth}
    \begin{prooftree}
      \AXC{$\catLF{X / Y}{f}$}
      \AXC{$\catLF{Y}{a}$}
      \RL{$>$}
      \BIC{$\catLF{X}{f\,a}$}
    \end{prooftree}
  \end{minipage}

  \item Backward functional application rule \\
  \begin{minipage}[h]{\linewidth}
    \begin{prooftree}
      \AXC{$\catLF{Y}{a}$}
      \AXC{$\catLF{X \bs Y}{f}$}
      \RL{$<$}
      \BIC{$\catLF{X}{f\,a}$}
    \end{prooftree}
  \end{minipage}

  \item Forward functional composition rule\\
  \begin{minipage}[h]{\linewidth}
    \begin{prooftree}
      \AXC{$\catLF{X / Y}{f}$}
      \AXC{$\catLF{Y / Z}{g}$}
      \RL{$\fcf$}
      \BIC{$\catLF{X / Z}{\lambda x. f(g\,x)}$}
    \end{prooftree}
  \end{minipage}

  \item Backward functional composition rule\\
  \begin{minipage}[h]{\linewidth}
    \begin{prooftree}
      \AXC{$\catLF{Y \bs Z}{g}$}
      \AXC{$\catLF{X \bs Y}{f}$}
      \RL{$\fcb$}
      \BIC{$\catLF{X \bs Z}{\lambda x. f(g\,x)}$}
    \end{prooftree}
  \end{minipage}
  
  \item Forward functional crossed composition rule\\
  \begin{minipage}[h]{\linewidth}
    \begin{prooftree}
      \AXC{$\catLF{X / Y}{f}$}
      \AXC{$\catLF{Y \bs Z}{g}$}
      \RL{$\cfcf$}
      \BIC{$\catLF{X \bs Z}{\lambda x. f(g\,x)}$}
    \end{prooftree}
  \end{minipage}

  \item Backward functional crossed composition rule\\
  \begin{minipage}[h]{\linewidth}
    \begin{prooftree}
      \AXC{$\catLF{Y / Z}{g}$}
      \AXC{$\catLF{X \bs Y}{f}$}
      \RL{$\cfcf$}
      \BIC{$\catLF{X / Z}{\lambda x. f(g\,x)}$}
    \end{prooftree}
  \end{minipage}

\end{itemize}

\section{Details of Axioms}
\label{apd:axioms}
\begin{table*}[h]
  \centering
  \begin{tabular}{cc}
  \hline
    Name & Logical Formula \\
  \hline
        CP &
      $\forall x\ y. \left( \left( \exists d. \left( \bA(x, d) \land \lnot \bA(y, d) \right) \right) \rightarrow \forall d. \left( \bA(y, d) \rightarrow \bA(x, d) \right) \right)$\\
  \hline \hline
      ANT &
      $\forall x\ d. \left( \bP(x, d) \leftrightarrow \lnot \bN(x, d) \right)$\\
  \hline \hline
      UP &
      $\forall x\ d. \left( \bP(x, d) \rightarrow \forall d'. \left( d' \leq d \rightarrow \bP(x, d') \right) \right)$ \\
      DOWN &
      $\forall x\ d. \left( \bN(x, d) \rightarrow \forall d'. \left( d' \geq d \rightarrow \bN(x, d') \right) \right)$ \\
  \hline \hline
      DELTA &
      $\forall x. \left( \bA(x, \theta - \delta) \leftrightarrow \bA(x, \theta + \delta) \right)$ \\
  \hline
  \end{tabular}
  \caption{Axioms for Japanese comparatives. $\bA$ denotes adjectives, $\bP$ denotes positive adjectives such as \textit{long}, and $\bN$ denotes negative adjectives such as \textit{short}.}
  \label{tab:axioms}
\end{table*}

\Cref{tab:axioms} shows the axioms employed in our system.
(CP) is the axiom we already introduced in \Cref{subsec:theorem-proving}.
We can make the following inferences using this axiom.
(\ref{cp-1}) and (\ref{cp-2}) are the premises, and (\ref{cp-3}) is the hypothesis.

\begin{exe}
    \ex\label{ex:consistency_postulate_inference}
    \begin{xlist}
        \ex\label{cp-1}
            \gll Taro-wa Jiro yori omoi.\\
                Taro-\Top{} Jiro than heavy\\
            \glt ``Taro is heavier than Jiro.''
        \ex\label{cp-2}
            \gll Jiro-wa 70 kg yori omoi.\\
                Jiro-\Top{} 70 kg than heavy\\
            \glt ``Jiro is heavier than 70 kg.''
        \ex\label{cp-3}
            \gll Taro-wa 70 kg yori omoi.\\
                Taro-\Top{} 70 kg than heavy\\
            \glt ``Taro is heavier than 70 kg.''
        \sn\hfill(entailment)
    \end{xlist}
\end{exe}

\noindent
Concretely, from (CP) and (\ref{cp-1}), we obtain $\heavy(\jiro, 70) \rightarrow \heavy(\taro, 70)$.
Then, from this formula and (\ref{cp-2}), we can derive $\heavy(\taro, 70)$.

(ANT) indicates the antonymy relation between positive and negative adjectives.
The following is an example of an inference using this axiom.
(\ref{ant-1}) is the premise and (\ref{ant-2}) is the hypothesis.
\begin{exe}
    \ex\label{ex:antonymy_inference}
    \begin{xlist}
        \ex\label{ant-1}
            \gll Taro-wa Jiro yori omoi.\\
                Taro-\Top{} Jiro than heavy\\
            \glt ``Taro is heavier than Jiro.''
        \ex\label{ant-2}
            \gll Taro-wa Jiro yori karui.\\
                Taro-\Top{} Jiro than light\\
            \glt ``Taro is lighter than Jiro.''
        \sn\hfill(contradiction)
    \end{xlist}
\end{exe}

(UP) and (DOWN) are axioms that indicate the monotonicity of positive and negative adjectives, respectively.
(DELTA) is an axiom about equatives.
Using these axioms, we can prove the entailment relation in (\ref{ex:japanese_equative_inference}) as follows.
(\ref{jeis-1}), (\ref{jeis-2}), and (\ref{jeis-3}) are the semantic representations of (\ref{jei-1}), (\ref{jei-2}), and (\ref{jei-3}), respectively.
\begin{exe}
    \ex\label{ex:japanese_equative_inference_semantic}
    \begin{xlist}
        \ex\label{jeis-1}
        $\begin{aligned}[t]
        &\forall d_1\,d_2. \left( \left( \lnot\left( \heavy(\taro, d_1) \right.\right.\right. \\
        &{}\leftrightarrow \left.\heavy(\jiro, d_1) \right)\\
        &{}\land \lnot\left( \heavy(\taro, d_2)\right.\\
        &{}\leftrightarrow \left.\left.\heavy(\jiro, d_2) \right) \right) \\
        &{}\rightarrow \left.| d_1 - d_2 | < \delta \right)
        \end{aligned}$
        \ex\label{jeis-2}
        $\heavy(\jiro, \theta)$
        \ex\label{jeis-3}
        $\heavy(\taro, \theta)$
    \end{xlist}
\end{exe}
First, from (\ref{jeis-2}), (UP), and (DELTA), we can derive $\heavy(\jiro, \theta)$ and $\heavy(\jiro, \theta + \delta)$.
Then, by replacing $d_1$ (resp.\ $d_2$) in (27a) with $\theta+\delta$ (resp.\ $\delta$), and by contraposition, we obtain either $\heavy(\taro, \theta + \delta) \leftrightarrow \heavy(\jiro, \theta + \delta)$ or $\heavy(\taro, \theta) \leftrightarrow \heavy(\jiro, \theta)$.
In both cases, $\heavy(\taro, \theta)$ is true since we have $\heavy(\jiro, \theta + \delta)$ and $\heavy(\jiro, \theta)$.

In the implementation, (CP) and (DELTA) are added for all gradable adjectives.
(ANT) is added for adjectives that have an antonym.
(UP) and (DOWN) are added for positive adjectives and negative adjectives, respectively.

\section{Problem Replacement}

\Cref{tab:replaced-example} shows an example of the problems in JSeM to which we applied the replacement we discussed in~\cref{sec:dataset}.
The original problem uses the property that the presupposition ``Hanako is an early riser'' is not affected by the modal ``kamo-sire-nai.''
We did not implement the semantic representation of modals, so we replaced them with a negation ``to-iu-wake-de-wa-nai.''
Since presuppositions are unaffected by negation (as well as by modals), this replacement does not alter the purpose of the problem---namely, to test whether the model understands that presuppositions are not influenced by entailment-canceling operators.
\begin{table*}[h]
  \centering
  \begin{tabular}{ll}
    \hline
    \multicolumn{2}{l}{jsem-621 (\textbf{original}), Gold answer: yes} \\
    \hline \hline
    Premise &
    \begin{tabular}[c]{@{}l@{}}
      Taro-wa Hanako izyoo-ni hayaoki kamo-sire-nai.\\
      (Taro may be an earlier riser than Hanako.)
    \end{tabular}
    \\
    Hypothesis &
    \begin{tabular}[c]{@{}l@{}}
      Hanako-wa Hayaoki-da.\\
      (Hanako is an early riser.)
    \end{tabular}
    \\
    \hline
    \multicolumn{2}{l}{jsem-621 (\textbf{replaced}), Gold answer: yes} \\
    \hline \hline
    Premise &
    \begin{tabular}[c]{@{}l@{}}
      Taro-wa Hanako izyoo-ni hayaoki toiu-wake-de-wa-nai.\\
      (Taro is not an earlier riser than Hanako.)
    \end{tabular}
    \\
    Hypothesis &
    \begin{tabular}[c]{@{}l@{}}
      Hanako-wa Hayaoki-da.\\
      (Hanako is an early riser.)
    \end{tabular}
    \\
    \hline
  \end{tabular}
  \caption{An example of the problems in which we replaced a modal with a negation}
  \label{tab:replaced-example}
\end{table*}

\section{Prompts for the Baseline Models}
\label{apd:prompts}
\begin{table*}[ht]
  \centering
  \begin{tabular}{cl}
    \hline
    system &
    \begin{tabular}[c]{@{}l@{}}
      前提文と仮説文が与えられます。 \\
      前提文が仮説文を含意しているか答えてください。 \\
      「含意」、「矛盾」、「中立」のいずれかで答えてください。\\
      (You are given premises and a hypothesis. \\
      Answer whether the premises entail the hypothesis.\\
      Answer with ``entailment'', ``contradiction'', or ``neutral.'')
    \end{tabular}
    \\
    \hline
    user &
    \begin{tabular}{l}
      前提１：PC-6082はITEL-XZより速い。\\
      前提２：ITEL-XZは速い。\\
      仮説：PC-6082は速い。\\
      (Premise 1: PC-6082 is faster than ITEL-XZ. \\
      Premise 2: ITEL-XZ is fast. \\
      Hypothesis: PC-6082 is fast.)
    \end{tabular}
    \\
    \hline
  \end{tabular}
  \caption{Example of the prompt for GPT-4o}
  \label{tab:gpt-prompt}
\end{table*}

\begin{table*}
    \centering
    \begin{tabular}{l}
    \hline
        前提文と仮説文が与えられます。\\
        前提文が仮説文を含意しているか答えてください。\\
        「含意」、「矛盾」、「中立」のいずれかで答えてください。\\
        前提１：PC-6082はITEL-XZより速い。\\
        前提２：ITEL-XZは速い。\\
        仮説：PC-6082は速い。\\
        回答：\\
    \hline
    \end{tabular}
    \caption{Example of the prompt for Swallow}
    \label{tab:swallow-prompt}
\end{table*}

\Cref{tab:gpt-prompt} and \Cref{tab:swallow-prompt} show examples of prompts for GPT-4o and Swallow, respectively.

\clearpage
\section{Detailed Results}
\label{apd:results}
\begin{table*}[h]
    \centering
    \begin{tabular}{cccccc}
    \hline
        Prompt Type & Accuracy & Gold Label & Precision & Recall & F1 Score  \\
    \hline\hline
        \multirow{3}{*}{E-C-N} & \multirow{3}{*}{0.619} & E & 0.62 & 1.00 & 0.76 \\
        & & C & 0.00 & 0.00 & 0.00 \\
        & & N & 0.67 & 0.10 & 0.17 \\
    \hline
        \multirow{3}{*}{E-N-C} & \multirow{3}{*}{0.605} & E & 0.61 & 1.00 & 0.76 \\
        & & C & 0.50 & 0.12 & 0.20 \\
        & & N & 0.00 & 0.00 & 0.00 \\
    \hline
        \multirow{3}{*}{C-E-N} & \multirow{3}{*}{0.225} & E & 0.80 & 0.19 & 0.31 \\
        & & C & 0.13 & 1.00 & 0.23 \\
        & & N & 0.00 & 0.00 & 0.00 \\
    \hline
        \multirow{3}{*}{C-N-E} & \multirow{3}{*}{0.591} & E & 0.77 & 0.81 & 0.79 \\
        & & C & 0.30 & 1.00 & 0.46 \\
        & & N & 0.00 & 0.00 & 0.00 \\
    \hline
        \multirow{3}{*}{N-E-C} & \multirow{3}{*}{0.605} & E & 0.60 & 1.00 & 0.75 \\
        & & C & 1.00 & 0.12 & 0.22 \\
        & & N & 0.00 & 0.00 & 0.00 \\
    \hline
        \multirow{3}{*}{N-C-E} & \multirow{3}{*}{0.647} & E & 0.64 & 1.00 & 0.78 \\
        & & C & 0.75 & 0.38 & 0.50 \\
        & & N & 1.00 & 0.05 & 0.09 \\
    \hline
    \end{tabular}
    \caption{Evaluation results of Swallow 8B on each prompt}
    \label{tab:results_swallow-8B}
\end{table*}

\begin{table*}[h]
    \centering
    \begin{tabular}{cccccc}
    \hline
        Prompt Type & Accuracy & Gold Label & Precision & Recall & F1 Score  \\
    \hline\hline
        \multirow{3}{*}{E-C-N} & \multirow{3}{*}{0.647} & E & 0.80 & 0.86 & 0.83 \\
        & & C & 0.36 & 1.00 & 0.53 \\
        & & N & 0.50 & 0.10 & 0.16 \\
    \hline
        \multirow{3}{*}{E-N-C} & \multirow{3}{*}{0.690} & E & 0.81 & 0.83 & 0.82 \\
        & & C & 0.55 & 0.75 & 0.63 \\
        & & N & 0.47 & 0.38 & 0.42 \\
    \hline
        \multirow{3}{*}{C-E-N} & \multirow{3}{*}{0.676} & E & 0.80 & 0.86 & 0.83 \\
        & & C & 0.40 & 1.00 & 0.57 \\
        & & N & 0.67 & 0.19 & 0.30 \\
    \hline
        \multirow{3}{*}{C-N-E} & \multirow{3}{*}{0.661} & E & 0.80 & 0.86 & 0.83 \\
        & & C & 0.42 & 1.00 & 0.59 \\
        & & N & 0.43 & 0.14 & 0.21 \\
    \hline
        \multirow{3}{*}{N-E-C} & \multirow{3}{*}{0.760} & E & 0.88 & 0.83 & 0.85 \\
        & & C & 0.62 & 1.00 & 0.76 \\
        & & N & 0.61 & 0.52 & 0.56 \\
    \hline
        \multirow{3}{*}{N-C-E} & \multirow{3}{*}{0.788} & E & 0.88 & 0.86 & 0.87 \\
        & & C & 0.62 & 1.00 & 0.76 \\
        & & N & 0.71 & 0.57 & 0.63 \\
    \hline
    \end{tabular}
    \caption{Evaluation results of Swallow 70B on each prompt}
    \label{tab:results_swallow-70B}
\end{table*}

\begin{table*}[h]
    \centering
    \begin{tabular}{cccccc}
    \hline
        Prompt Type & Accuracy & Gold Label & Precision & Recall & F1 Score  \\
    \hline\hline
        \multirow{3}{*}{E-C-N} & \multirow{3}{*}{0.746} & E & 0.83 & 0.81 & 0.82 \\
        & & C & 0.75 & 0.75 & 0.75 \\
        & & N & 0.59 & 0.62 & 0.60 \\
    \hline
        \multirow{3}{*}{E-N-C} & \multirow{3}{*}{0.774} & E & 0.84 & 0.86 & 0.85 \\
        & & C & 0.75 & 0.75 & 0.75 \\
        & & N & 0.65 & 0.62 & 0.63 \\
    \hline
        \multirow{3}{*}{C-E-N} & \multirow{3}{*}{0.774} & E & 0.85 & 0.83 & 0.84 \\
        & & C & 0.78 & 0.88 & 0.82 \\
        & & N & 0.62 & 0.62 & 0.62 \\
    \hline
        \multirow{3}{*}{C-N-E} & \multirow{3}{*}{0.760} & E & 0.85 & 0.81 & 0.83 \\
        & & C & 0.78 & 0.88 & 0.82 \\
        & & N & 0.59 & 0.62 & 0.60 \\
    \hline
        \multirow{3}{*}{N-E-C} & \multirow{3}{*}{0.788} & E & 0.86 & 0.86 & 0.86 \\
        & & C & 0.75 & 0.75 & 0.75 \\
        & & N & 0.67 & 0.67 & 0.67 \\
    \hline
        \multirow{3}{*}{N-C-E} & \multirow{3}{*}{0.788} & E & 0.86 & 0.86 & 0.86 \\
        & & C & 0.78 & 0.88 & 0.82 \\
        & & N & 0.65 & 0.62 & 0.63 \\
    \hline
    \end{tabular}
    \caption{Evaluation results of GPT-4o on each prompt}
    \label{tab:results_gpt-4o}
\end{table*}

\Cref{tab:results_swallow-8B}, \Cref{tab:results_swallow-70B}, and \Cref{tab:results_gpt-4o} show the detailed evaluation results of each baseline model.
E, C, and N represent entailment, contradiction, and neutral, respectively.
``Prompt Type'' indicates the order of the words 含意 (entailment), 矛盾 (contradiction), and 中立 (neutral) as they appear in each prompt.
For example, \Cref{tab:gpt-prompt} and \Cref{tab:swallow-prompt} show prompts of the E-C-N (含意-矛盾-中立) type.

Swallow 8B tended to output \textit{yes} when 含意 or 中立 appeared first in the prompt, resulting in substantially lower F1 scores for contradiction and neutral compared to entailment.
Conversely, when 矛盾 was presented first, the number of \textit{no} responses increased.

In contrast, Swallow 70B and GPT-4o produced more balanced outputs, achieving higher F1 scores than Swallow 8B.

\end{document}